\documentclass[journal]{IEEEtran}
\usepackage{amsmath,amsfonts}
\usepackage{algorithmic}
\usepackage{algorithm}
\usepackage{array}
\usepackage[caption=false,font=normalsize,labelfont=sf,textfont=sf]{subfig}
\usepackage{textcomp}
\usepackage{stfloats}
\usepackage{url}
\usepackage{verbatim}
\usepackage{graphicx}
\usepackage{cite}
\usepackage{multirow}
\usepackage{amssymb}

\usepackage{booktabs}

\begin{document}

\title{Scattering-induced entropy boost for highly-compressed optical sensing and encryption}
\author{Xinrui Zhan, Xuyang Chang, Daoyu Li, Rong Yan, Yinuo Zhang, Liheng Bian$^{*}$
\thanks{Xinrui Zhan, Xuyang Chang, Daoyu Li, Rong Yan, Yinuo Zhang and Liheng Bian are with MIIT Key Laboratory of Complex-field Intelligent Sensing, Beijing Institute of Technology, Beijing 100081, China}
\thanks{Corresponding author: Liheng Bian (bian@bit.edu.cn)}
}


\maketitle

\begin{abstract}
Image sensing often relies on a high-quality machine vision system with a large field of view and high resolution. It requires fine imaging optics, has high computational costs, and requires a large communication bandwidth between image sensors and computing units. In this paper, we propose a novel image-free sensing framework for resource-efficient image classification, where the required number of measurements can be reduced by up to two orders of magnitude. In the proposed framework for single-pixel detection, the optical field for a target is first scattered by an optical diffuser and then two-dimensionally modulated by a spatial light modulator. The optical diffuser simultaneously serves as a compressor and an encryptor for the target information, effectively narrowing the field of view and improving the system's security. The one-dimensional sequence of intensity values, which is measured with time-varying patterns on the spatial light modulator, is then used to extract semantic information based on end-to-end deep learning. The proposed sensing framework is shown to obtain over a 95\% accuracy at sampling rates of 1\% and 5\% for classification on the MNIST dataset and the recognition of Chinese license plates, respectively, and the framework is up to 24\% more efficient than the approach without an optical diffuser. The proposed framework represents a significant breakthrough in high-throughput machine intelligence for scene analysis with low bandwidth, low costs, and strong encryption.
\end{abstract}


\section{Introduction}
\IEEEPARstart{W}{ith} the development of deep learning techniques and computational hardware platforms, machine vision, in place of human knowledge, has brought about high-level semantic understanding in multiple fields, such as unmanned driving, the automation industry, and space engineering \cite{snyder2004machine}. 
To make decisions with high accuracy, current machine vision techniques require large-scale and high-resolution images, leading to high computational complexity and considerable requirements for imaging devices and computing resources\cite{jordan2015machine}. For certain resource-limited platforms (such as drones and micronano satellites) with insufficient data, a narrow bandwidth, or a low hashrate, the existing machine vision techniques that rely on large-scale input and processing are not applicable. Lightweight techniques for efficient sensing are more suitable and desirable for such tasks \cite{mennel2020ultrafast}.

As machine vision has developed, a turning point occurred at the beginning of the 21st century when compressive sensing was proposed \cite{candes2006compressive}. The compressive theory breaks the laws of the Nyquist sampling limit and allows signals to be reconstructed based on a small amount of collected data, thus attracting much attention for its high security as well as its low bandwidth in data storage and transmission. Reconstruction algorithms based on sparse data usually employ an iterative strategy, such as the conjugate gradient descent and alternating projection strategies \cite{edgar2019principles, watts2014terahertz, duarte2008single}. The emergence and prosperity of deep learning have further improved the reconstruction quality and efficiency, especially at low sampling rates \cite{zhang2015single, sun20133d}.
However, the high-precision reconstructed results usually contain ineffectual information, such as background information, and details that are useless for making decisions. Moreover, the above techniques still require an additional step to extract high-level semantic information, which is redundant and imposes a heavy burden on computing resources.

In this work, we present a novel lightweight image-free technique that can directly extract multitarget semantic information from highly compressed measurements. Moreover, for the first time, we introduce proactive scattering for light modulation by utilizing the information entropy boost induced by scattering for high-precision sensing and high-security encryption. In comparison to the majority of the current optical approaches, this innovation effectively increases the decoding efficiency in terms of both the data amount and sensing accuracy.

\begin{figure*}[h]
\centering
\includegraphics[width=1.0\linewidth]{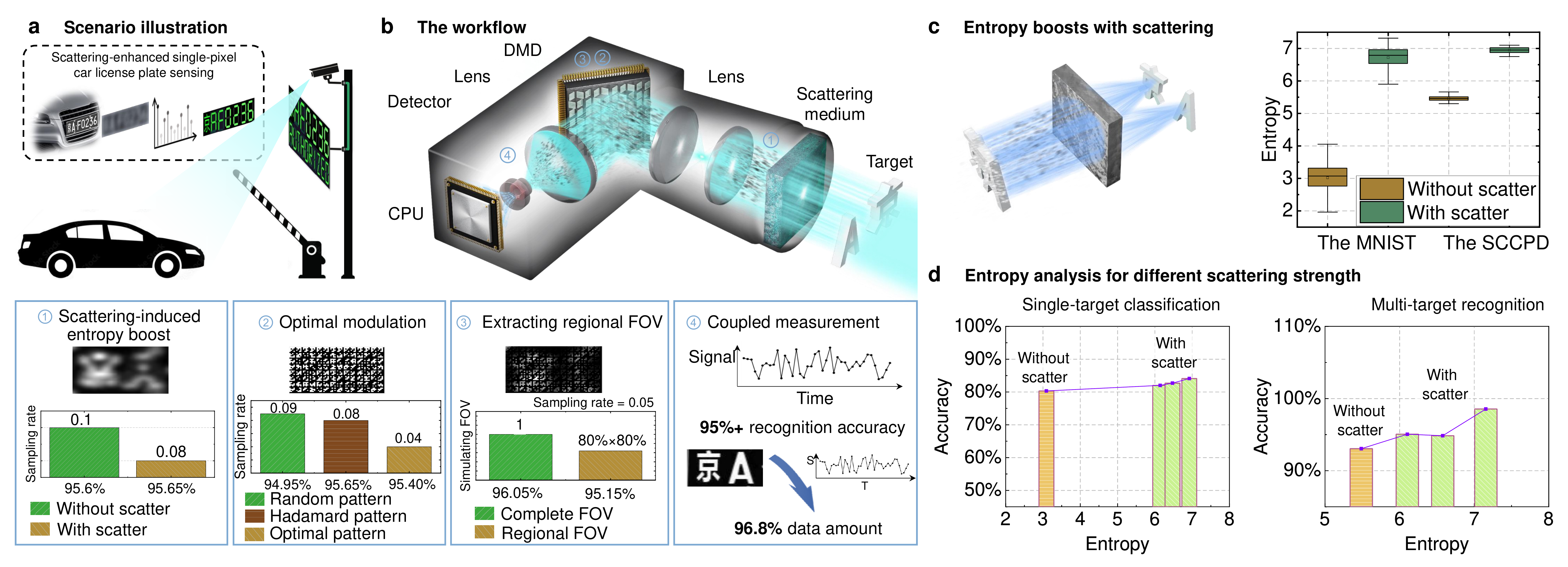}
\caption{The framework of the reported method.
(a) A scenario illustration in car plate license sensing. (b) Optical transformations in the workflow. (c) Entropy boost induced by scattering. We use a light-like image to symbolize the transformation of information. The target's different areas of information are mixed into a section after scattering, which produces a significant entropy boost. Therefore, we propose to only extract the regional field of view (FOV) of the scattering image for optical sensing and encryption. The left figure presents the quantitative analysis of the scattering-induced entropy boosts by calculating the average entropy of each image dataset. (d) Analysis of different scattering strengths. The `entropy' on the x-axis represents the average entropy of the training images with different scattering intensities, and the `accuracy' on the y-axis is the sensing accuracy on the test dataset at the same sampling rate with a regional FOV. For single-pixel sensing, we select only a 50\% field width at a 0.05 sampling rate. For multitarget recognition, the corresponding parameters are set to 8\% and 0.1.}
\label{fig_equipment}
\end{figure*}

The reported technique has the same origin as single-pixel imaging\cite{edgar2019principles}, compressed learning \cite{zisselman2018compressed}, and speckle-encoded imaging \cite{ruan2020fluorescence}. 
Unlike single-pixel imaging, this technique works by direct inference from a one-dimensional coupled measurement based on an efficient decoding network, avoiding the process of image reconstruction that contributes to a notable reduction in computation speed \cite{lohit2015reconstruction, fu2020single}. 
We report an end-to-end image-free technique that is different from the existing compressed learning techniques. Not only do we achieve joint optimization for both pattern modulation and target decoding, but we also implement it for multitarget recognition on more complicated and information-disrupting car license plate datasets rather than for simple classification tasks. Moreover, the reported technique further promotes efficiency by introducing scattering to enhance the information entropy. In recent studies, scattering has also been applied as the medium for lensless imaging instead of using fine and delicate equipment\cite{judkewitz2013speckle, ruan2020fluorescence}.
However, unlike the existing scattering-introduced techniques that only adopt scattering for light modulation \cite{vellekoop2010exploiting, jang2018wavefront, ryu2016optical}, we utilize the intrinsic feature of scattering, that is, the information multiplexing process, and then we accomplish information coupling combined with pattern modulation to achieve further data reduction with high-precision perception.

In theory, according to the Shannon formula, the information entropy of an image $I$ can be expressed as follows \cite{wu2013local}:
\begin{equation}
H\left(\mathbf{I}\right) = \sum_{l=1}^{L} P(l) \log \frac{1}{P(l)},
\end{equation}
where $L$ is the number of pixel value ranges, which is 256 for a grayscale image, and $P(l)$ represents the probability at the $l$th pixel value. When an image is completely random, its pixels are evenly distributed, and it reaches the maximum information entropy, where $P(l) = 1 / 256, H(I) = 8$. According to this formula, we calculate the information entropy for several images and their corresponding images with scattering, as shown in Fig. \ref{fig_equipment} (c). The average information entropy of the Modified National Institute of Standards and Technology (MNIST) dataset images without scattering is 3.092. After scattering, the information entropy of the same scene will increase to 6.773, which is approximately 2 times higher. Note that in this work, we assume that an information entropy boost leads to the expansion of valuable information, which will be validated in the experiments presented below. The information entropy boost means that scattering multiplexes global information into regional fields, thus achieving a compact coupling of feature information \cite{Lyu2019Learning}. For sensing tasks, the information entropy boost can be referred to as an increase in the randomness of a detected measurement, which reveals an expansion of the key space; thus, there is a great possibility of extracting feature information and realizing high-precision sensing. For optical encryption tasks, a high information entropy indicates a random state of ciphertext and a small probability of information leakage, which is of great practical potential in the cryptography field.

In the following experiments, we explore the positive effect of scattering on both single-pixel classification and multitarget recognition for extremely small amounts of data with regional detected areas, which enables us to overcome the fundamental deficiencies in traditional large-scale methods, with a significant reduction in computation cost and storage burden. Specifically, for single-target classification, we obtain an accuracy of 95.89\% on the MNIST dataset at a low sampling rate of 1.25\% with only a quarter of the original vision fields and achieved 84.09\% on Fashion-MNIST with the same sampling mode, which is 4\% higher than that in the same vision field without scattering. For multitarget recognition, a high accuracy of 97.40\% can be obtained on the Simulated Chinese Car Plate dataset (SCCPD) at a 6.4\% sampling rate with a width scalability of 80\%, which is, notably, 24\% better than the approach without scattering when using the same amount of data. The above experiments support the efficiency of our reported technique for high-precision optical sensing and encryption. Moreover, we also analyze the cryptography security with the National Institute of Standards and Technology (NIST) test. The result shows that compared with the traditional single-pixel encryption method \cite{clemente2010optical} that only adopts pattern modulation, the reported technique quantitively enhances the randomness of the ciphertext by approximately two times, which notably promotes encryption security.

\section{Methods}

\subsection{The reported scattering-induced image-free sensing and optical encryption.}

In Fig. \ref{fig_equipment}(b), we depict the workflow of the scattering-induced optical sensing and encryption method. The target light field is first scattered by a diffuser, which enables the multiplexing coupling of target information. A set of lenses is arranged to resize the target light to an appropriate FOV. Then, we employ a digital micromirror device (DMD) and a single-pixel detector to map the two-dimensional image to a one-dimensional coupled measurement. Additionally, with the information entropy boost induced by scattering, we only need to modulate a regional FOV to achieve further data reduction. We construct a training dataset that consists of highly compressed scattered data and its semantic classification label. Then, we employ a novel image-free sensing network, namely, the semantic decoding network, to learn the mapping relationship. Therefore, we can realize high-precision sensing with highly compressed measurements. 

\subsection{Scattering procedure}

According to the inherent optical characteristics of photon propagation in scattering media and to verify the universality and reliability of the reported method, we use the following two simulation strategies to generate scattering images for single-target and multitarget recognition.

\textbf{Speckle image simulation based on the Monte Carlo model \cite{bar2019a}}: The scattering process can be described as a series of random events conforming to a multivariate Gaussian distribution. According to the scattering parameters characterizing the statistical distribution of dielectric scatterers, the covariance space path equation of the scattering process is established. Therefore, the Monte Carlo scattering simulation method accurately calculates the speckle field to simulate the pattern. 

\textbf{Speckle images simulation based on constructing translation-invariant operators (ScatNet) \cite{mallat2012group}}: Based on the definition of the scattering propagator, namely, the path ordered product of nonlinear and noncommuting operators, each of which computes the modulus of a wavelet transform, Mallat \emph{et al.} constructed translation-invariant operators that are Lipschitz-continuous to the action of diffeomorphisms, upon which the scattering light field with translation invariance and rotation invariance can be accurately calculated. Hence, the scattered image can be obtained.

\subsection{Structural light illumination}

Structural modulation patterns are a set of two-dimensional matrices, the size of which is consistent with the selected field of view. In this work, we employ Hadamard patterns for modulation. Compared with other types of patterns, Hadamard illumination modulation performs better in converging spatial information in the transform domain because it forms a complete orthogonal set \cite{pratt1969hadamard}. The Hadamard pattern series is generated by the Hadamard matrix, which meets the following conditions:
\begin{eqnarray}
\label{eqs:e4}
\begin{gathered}
\mathbf{H} \mathbf{H}^{T}=\mathbf{H}^{T} \mathbf{H}=\mathbf{I}, \\
\mathbf{H}_{1} \triangleq \frac{1}{\sqrt{2}}\left[\begin{array}{rr}
1 & 1 \\
1 & -1
\end{array}\right], \\
\mathbf{H}_{n}=\mathbf{H}_{1} \otimes \mathbf{H}_{n-1}=\frac{1}{\sqrt{2}}\left[\begin{array}{rr}
\mathbf{H}_{n-1} & \mathbf{H}_{n-1} \\
\mathbf{H}_{n-1} & -\mathbf{H}_{n-1}
\end{array}\right],
\end{gathered}
\end{eqnarray}
where $H$ is the Hadamard matrix, $I$ is the identity matrix, and $n$ is an integer. In this work, we randomly exchanged the rows and columns of the Hadamard matrix H to obtain adequate modulation patterns. To simplify the optical experiment for negative pattern operation, we map all the parameters that have a value of -1 in the matrix to 0. 
Moreover, benefit from the information entropy boost induced by scattering, we can modulate only part of the light field to further reduce the number of measurements. Fig. \ref{fig_result_single} (c) depicts two different strategies in pattern design for regional FOV modulation. On the data acquisition end, a single-pixel detector id used to capture light intensity. By adapting the frequency of modulating patterns and the acquisition end, a series of one-dimensional single-pixel measurement will be collected and sent for feature extraction.

\subsection{Sensing network}

Image-free sensing network is employed to extract semantic information directly from the collected single-pixel measurement, and it also functions as the semantic decoding network in our optical encryption experiment.
For the single-target sensing task, we designed an image-free classification network based on the EfficientNet architecture \cite{tan2019efficientnet}, which is a sensing network constructed with a multidimensional hybrid model scaling method. EfficientNet requires a neural architecture search strategy to balance the three dimensions of the network, depth, width, and resolution, thus achieving a much better accuracy and efficiency than has been achieved by previous convolutional networks. For the Chinese car license plate sensing task, we designed a convolutional recurrent neural network (CRNN) model \cite{graves2006connectionist} to achieve unsegmented multitarget image-free recognition. The architecture of the CRNN is composed of a convolutional neural network (CNN), gated recurrent unit (GRU), and connectionist temporal classification (CTC) loss function. GRUs have been extensively employed for multitarget unsegmented sensing tasks, especially for the recognition of handwriting and speech. They overcome the deficiencies in recurrent neural networks (RNNs), such as long-term memory and gradient disappearance during backpropagation, thus achieving better performance and a significant reduction in the computational cost. The detailed network structures can be found in the attached supplementary material.

\subsection{Experimental setup} 

We built a proof-of-concept system to experimentally demonstrate the reported method. The scene was projected onto a screen by a projector (Panasonic x416c XGA). Then, the reflected light was scattered through the scattering media and modulated by a DMD (ViALUX V-7001) after converging through the lens, and the modulation patterns were the Hadamard pattern series reported in this work. After the modulated light converged through the lens, the coupling signal was collected by a single-pixel silicon photodetector (Thorlabs PDA100A2, 320-1100 nm). By adjusting the frequency of projection and DMD modulation, we detected a series of one-dimensional measurements with the single-pixel detector. Based on the highly compressed measurement, we were able to extract high-level semantic features using an image-free sensing network, which is illustrated in detail in our supplementary materials.

\section{Results}
\noindent \textbf{Single-target classification.} 

\begin{figure}[H]
\centering
\includegraphics[width=1.0\linewidth]{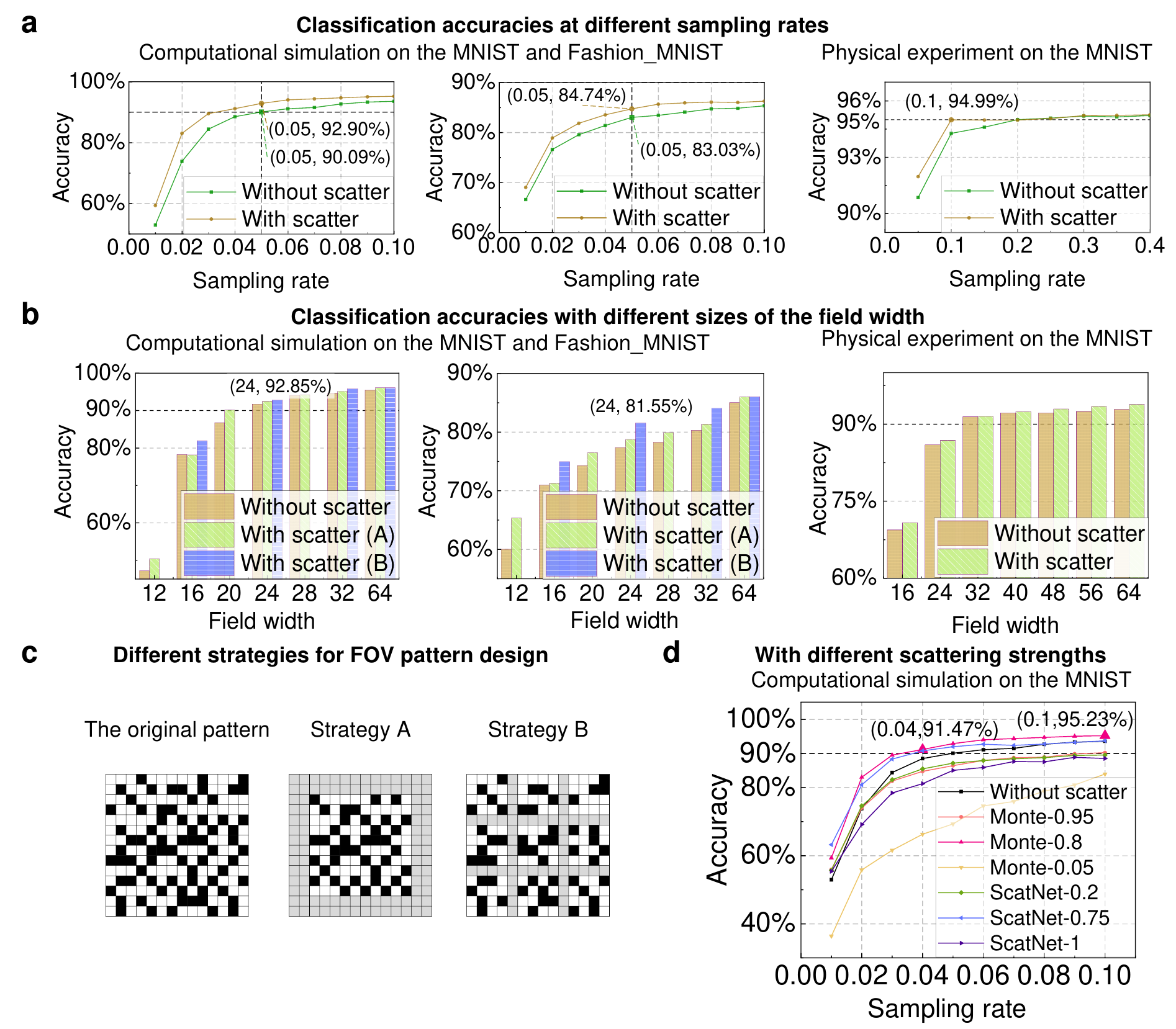}
\caption{Single-target image-free sensing results. (a) Sensing accuracy at different sampling rates, including computational simulations on the MNIST and Fashion-MNIST datasets (with `Monte-0.8' scattering intensity) and physical experiments on the MNIST dataset. (b) Sensing accuracy in the regional FOV.
In this work, the sampling rate is the ratio of the coupled measurement length to the size of the modulated field area. The term `field width' refers to the width of the modulated target area (the number of active columns or rows during modulation), which can be narrowed with a scattering-induced entropy boost. (c) Different strategies to extract the regional FOV data. The black blocks in the modulation patterns have values of zero where the light cannot pass through, and the white blocks have values of one. The gray line shows the adjustment that turns a one into a zero. (d) Sensing accuracy with different simulated scattering strengths on the MNIST dataset.}
\label{fig_result_single}
\end{figure}

The sampling rate in the single-pixel field is defined as the ratio of the length of the detected measurement, that is, the modulation pattern number, to the pixel number of the target view. To conclude, a higher sampling rate leads to more target information, which contributes to higher sensing accuracy but time-consuming detection and high-complexity computation. In this work, to analyze the influence of scattering on single-pixel sensing, we implement experiments on the MNIST\cite{lecun1998gradient} and Fashion-MNIST \cite{xiao2017fashion} datasets at different sampling rates from 0.01 to 0.1, as shown in Fig. \ref{fig_result_single} (a) and (b). The results validate that the sensing performance is promoted as the sampling rate increases regardless of the scattering effect. Similar to conventional single-pixel sensing without scattering, when the sampling rate is less than 0.05, the sensing accuracy of the measurement with scattering improves with increasing sampling rate, while the improvement speed of the sensing accuracy slows down or is almost constant when the sampling rate rises above 0.05. The turning point at 0.05 indicates an appropriate sampling rate that balances the sensing accuracy and acquisition complexity in practical applications. In the comparison of image-free sensing with and without scattering, the scattering enhancement is conspicuous at each sampling rate. Moreover, at the 0.05 sampling rate, the sensing result with scattering is approximately 2\% higher than the target without scattering. It is almost the same as the target without scattering at the 0.1 sampling rate on both the MNIST and Fashion-MNIST datasets, which indicates that the scattering enhancement strategy effectively reduces the sampling rate by half.

To further reduce the data amount, we utilize the information redundancy of scattering images and apply a high-precision sensing technique on only regional light fields. In the experiments presented below, considering that MNIST and Fashion-MNIST are both lightweight datasets containing 28$\times$28 images, we expand the image sizes to $64 \times 64$ to accentuate the impact of different regional vision fields on the sensing accuracy. As shown in Fig. \ref{fig_result_single}(c), we use two methods to extract partial vision fields by designing modulation patterns. The first approach is to only modulate the central part of the target (abbreviated as strategy `A'). The other strategy is similar to the linear interpolation algorithm in setting specific rows and columns as zero, and the rows and columns are evenly distributed in the two-dimensional pattern (abbreviated as strategy `B'). As shown in Fig. \ref{fig_result_single}(b), we set the sampling rate to 0.05, as discussed above, and the x-axis represents the width of the partial field or the number of unchanged rows and columns modulated in strategy `B.' 
When the field width is smaller than 20 pixels, the sensing accuracy varies noticeably with the fluctuation in field size, while it doesn't have much effect on the result when it exceeds 20 pixels. Therefore, it is concluded that 20 $\times$ 20 might be the equilibrium point for partial field sensing with scattering for both a small amount of data and acceptable performance. Compared with the method without scattering modulation in strategy `A,' we achieve better sensing performance at each sampling rate with scattering by adopting strategy `B' for modulation, especially on the Fashion-MNIST dataset, whose target is more information-disrupting at an extremely low sampling rate. Specifically, the data amount can be sharply reduced by 75\% with scattering, as the scattering-induced method achieves 84.09\% sensing accuracy on a 32$\times$32 regional field, which is approximately the same as that on the whole FOV (85.03\%). When comparing the two strategies for vision contraction, the sensing result with strategy `B' is better than that with strategy `A,' which verifies that the scattering light field with our reported technique multiplexes the image information into its adjacent region.

In practical applications, different types of scattering mediums have diverse performances in light modulation. Accordingly, the scattering images generated from the same scene differ substantially depending on the scattering medium. In this work, to analyze the impacts of different scattering strengths on the sensing results, we simulate and compare the image-free sensing results on the MNIST dataset with different scattering strengths, as shown in Fig. \ref{fig_result_single} (d), where `Monte-x' refers to the dataset generated based on the Monte Carlo model and `ScatNet-x' represents that generated based on ScatNet. In the `Monte-0.95,' `Monte-0.8,' and `Monte-0.05' datasets, the corresponding anisotropy constants are 0.95, 0.8, and 0.05, respectively. For the ScatNet modulation algorithm, the distinct scattering strength is simulated by adjusting the scattering particle density. The scattering strengths of `ScatNet-0.2,' `ScatNet-0.75,' and `ScatNet-1' are roughly reciprocal to the above Monte Carlo series with corresponding parameters set to 0.2, 0.75, and 1. An appropriate scattering strength of `Monte-0.8' and `ScatNet-0.75' apparently promotes better sensing accuracy than single-pixel sensing without scattering at each sampling rate, which verifies the effectiveness of the scattering enhancement. Specifically, at the most appropriate `Monte-0.8' scattering strength degree, we achieve higher than 90\% sensing accuracy when a 0.04 sampling rate is applied, and a 95.32\% precision value is obtained at a 0.1 sampling rate. Nevertheless, an excessively strong scattering strength might lead to precision attenuation.
Moreover, as shown in Fig. \ref{fig_equipment} (d), we exemplify the average information entropy of handwritten digits with three different scattering strengths simulated by the Monte Carlo algorithm. We adopt the same sampling rate of 0.1 and calculate the classification accuracy on the test data. The results reveal that higher information entropy induced by scattering contributes to richer feature information, which promotes the extraction ability for image-free sensing and confirms our former assumption in the introduction.

\noindent \textbf{Multitarget recognition.} 

Similarly, the reported framework can be applied to multitarget sensing. Although Xu \emph{et al.} \cite{xu2018towards} arranged a dataset composed of captured car plate images for detection and recognition, its sample distribution is unbalanced. A Chinese car license plate contains a Chinese character representing the province and six letters or numbers for identity authentication; however, more than 90\% of the data items in the dataset begin with the same Chinese character. We consider this dataset to be unable to influence the method's learning efficiency for complex Chinese characters. Therefore, we have implemented the reported image-free technique on a simulated Chinese car license plate with a uniformly distributed dataset.

\begin{figure}[h]
\centering
\includegraphics[width=1.0\linewidth]{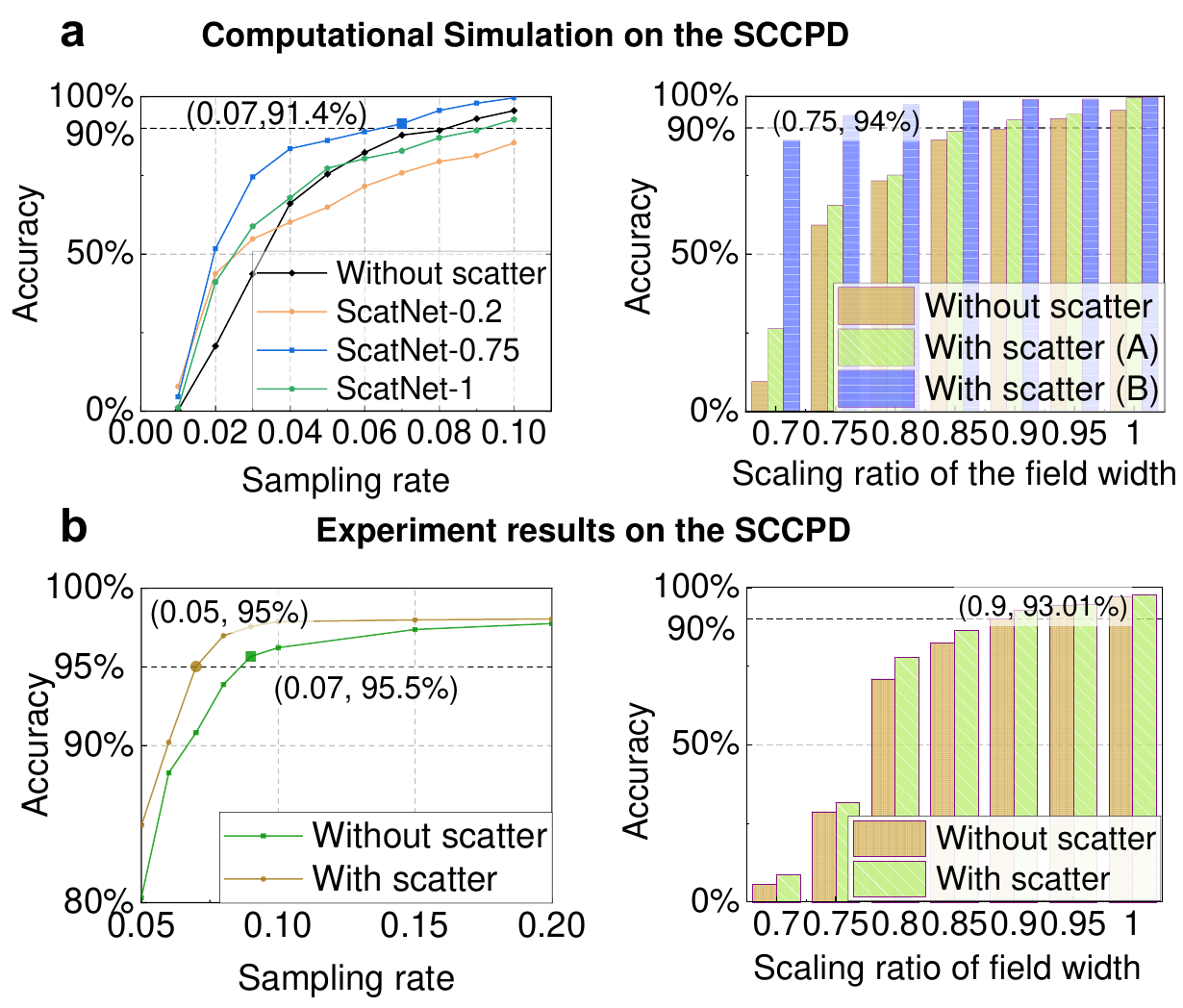}
\caption{Multitarget sensing results. (a) Simulation results on the SCCPD dataset with different scattering strengths at different sampling rates. (b) Experimental results on the SCCPD dataset at different sampling rates and different field widths.}
\label{fig_result_multi}
\end{figure}

The sensing accuracy in this part is defined as the complete string match accuracy. In the workflow of the Monte Carlo algorithm mentioned above, trouble may arise when dealing with a large target light field, as each photon will contact the tissue more than 100 times throughout the long propagation period. Therefore, we employ ScatNet, a lightweight technique, to computationally simulate the scattering process for the SCCPD dataset with different intensities. We adjust the scattering particle density to 0.2, 0.75, and 1, as illustrated in Fig. \ref{fig_result_multi} (a). 

When moderate scattering, such as that in `ScatNet-0.75,' is introduced in the optical path, the recognition accuracy on the SCCPD is substantially better than when modulating without scattering at each sampling rate, showing the effectiveness of the scattering-enhanced strategy for multitarget recognition. Moderate scattering considerably improves the recognition accuracy at each sampling rate, while excessive scattering may reduce the recognition performance of the system. Specifically, we achieve a sensing accuracy that is higher than 90\% when the sampling rate is 0.07. At a low sampling rate, such as rates less than 0.04, the SCCPD recognition accuracy for each scattering image is higher than it is without scattering, which enables highly compressed image-free sensing and demonstrates notable information enhancement induced by scattering.

In the next step, with the SCCPD dataset, we verify the effect of our reported method on multitarget recognition based on the partial FOV at a 0.1 sampling rate with moderate scattering strength. The acquisition method of the partial FOV is consistent with what is presented in Fig. \ref{fig_result_single} (b). As shown in Fig. \ref{fig_result_multi} (b), the abscissa represents the scaling ratio of the field width between the modulating area and the target vision. At the same sampling rate, the performance of multitarget image-free recognition based on the partial FOV of the scattering image is significantly better than that based on the partial FOV of the scene without scattering. Specifically, a high accuracy of 97.40\% can be obtained on the SCCPD at a 6.4\% sampling rate with a width scalability of 80\%, which is substantially better, that is, 24\% better than the accuracy with the original view using the same sampling mode without scattering. 

The experimental results for single-pixel sensing with `ScatNet-0.75' scattering intensity at a sampling rate of 0.1 are presented in Fig. \ref{fig_result_single} (a),(b) and Fig. \ref{fig_result_multi}(b). The MNIST images projected on the scene were scaled to 64 $\times$ 64, and the sizes of the SCCPD images were adjusted to 32 $\times$ 160. In both single-target and multitarget recognition tasks, the experimental results correspond with the simulations, validating the scattering-induced enhancement of sensing efficiency, especially at an extremely low sampling rate and in the regional FOV, which supports our reported technique on the compact multiplexing of feature information induced by scattering.

\noindent \textbf{Further investigations.}

To investigate the potential efficiency of the reported technique, we discussed end-to-end optimization with both modulating patterns and the decoding network for the best performance. The modulation procedure can be performed through a convolutional layer in the network, in which the modulation patterns act as the weight of the convolutional kernels. We implement end-to-end optimization on the same SCCPD dataset for training and obtain optimal patterns and well-trained network models, which can be generally applied to any Chinese car plate.

\begin{figure}[h]
\centering
\includegraphics[width=1\linewidth]{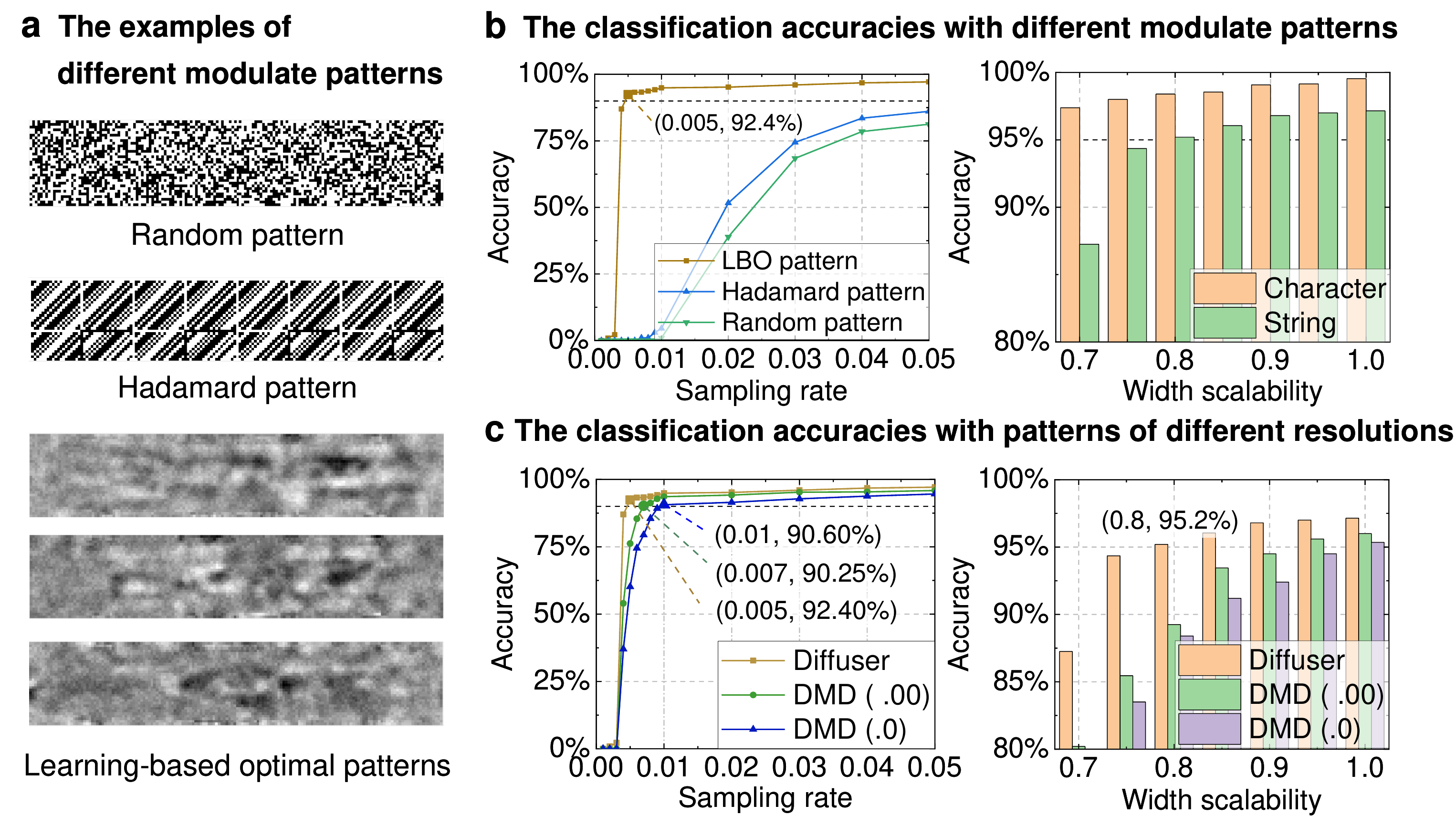}
\caption{Modulation sensing results for multitarget recognition. (a) Examples of random patterns and optimal patterns. (b) Sensing results with learning-based optimal patterns, where `LBO' represents the learning-based optimal and the `width scalability' is the scale ratio of the selected regional FOV to the original light width. (c) Sensing results when modulated patterns are replaced by the diffuser at different numerical precisions.}
\label{fig_add_result}
\end{figure}

As shown in Fig. \ref{fig_add_result}(a), the learning-based patterns are grayscale and contain image features learned from the training dataset. In Fig. \ref{fig_add_result}(b), we show the sensing results for both the single character recognition accuracy and string match accuracy with the `ScatNet(0.75)' scattering strength. Learning-based optimal patterns promote higher sensing accuracy, especially at lower sampling rates. At a 0.005 sampling rate, it achieves a 98.91\% accuracy on a single character match and 92.40\% on a string match, which is roughly the same high efficiency as when modulated by Hadamard patterns at a sampling rate of 0.02. Furthermore, we select regional areas at a 0.05 sampling rate, where we achieve almost the same high efficiency, indicating that compared with the Hadamard patterns, a 50\% reduction in the data amount is accomplished. However, unlike the optimal Hadamard patterns mentioned above, which are composed of fixed values of zero or one, the learning-based optimal pattern series should be trained separately when dealing with different scattering strengths. As analyzed above, if there are sufficient computing resources without any time limit, we suggest utilizing the optimal gray patterns for modulation for the best sensing efficiency.

We also explore the practicability of whether the scattering diffuser can be replaced by existing light equipment such as the DMD. Specifically, the scattering process can be simplified as a transfer matrix. Thus, we combine the scattering transfer matrix with the learning-based optimal patterns (as shown in Fig. \ref{fig_add_result}(a)) and project the mixed gray modulation patterns using the DMD. Despite the advantages of the DMD equipment price and modulation speed, the scattering diffuser is superior in its continuous light modulation effect. Therefore, we modulated the modulation patterns with different resolutions by reserving different decimals. As shown in Fig. \ref{fig_add_result}(c), the scattering introduced by the diffuser achieves better information coupling and higher sensing accuracy at each sampling rate. Therefore, the reported workflow cannot be modified when the physical scattering process cannot be replaced by light modulation equipment.

\noindent \textbf{Optical encryption.} 

\begin{table*}
\centering
\caption{NIST Test results.}
\label{tab_NIST}
\begin{tabular}{c|c|c|c|c}
\hline
\multirow{2}[4]{*}{NIST factor \textbackslash{} Target scene} & \multicolumn{2}{c|}{MNIST} & \multicolumn{2}{c}{Fashion-MNIST} \\
\cline{2-5} & Without scatter & With scatter & Without scatter & With scatter \\
\hline
Frequency & - & \textbf{0.19} & 0.24 & \textbf{0.89} \\
BlockFrequency & 1.00 & 1.00 & 1.00 & 1.00 \\
CumulativeSums & - & \textbf{0.24} & 0.21 & \textbf{0.92} \\
Rank & - & - & - & - \\
FFT & - & \textbf{0.12} & 0.59 & \textbf{0.65} \\
NonOverlappingTemplate & - & - & 0.22 & 0.18 \\
OverlappingTemplate & - & - & - & - \\
Universal & 0.1 & \textbf{0.38} & 0.45 & \textbf{0.81} \\
ApproximateEntropy & - & - & - & - \\
RandomExcursions & 0.51 & 0.04 & 0.12 & \textbf{0.13} \\
Serial & - & \textbf{0.06} & - & \textbf{0.11} \\
LinearComplexity & 0.6 & \textbf{0.83} & 0.56 & \textbf{0.99} \\
\bottomrule
\end{tabular}%
\end{table*}%
As analyzed above, the combination of single-pixel modulation and a scattering transformation achieves high-precision sensing on highly compressed measurements; hence, it can be applied to cryptography as well. That is, the target scene without scattering is a plaintext scene, and the ciphertext is a one-dimensional sequence modulated by scattering and single-pixel patterns. In the cryptography field, the NIST test \cite{pareschi2012statistical} is a widely used standard test for measuring the uncertainty of data. Generally, stronger uncertainty of ciphertext means better encryption performance, which could enhance the difficulty of deciphering ciphertext. The latest NIST test is composed of 15 testing items, including a frequency inspection and block frequency inspection. To simulate the encryption process of an actual scene, we expand a randomly selected MNIST image to 1000 $\times$ 1000 and simulate its scattering image of the same size, and then we employ a series of Hadamard patterns for modulation at a sampling rate of 1. 
Table \ref{tab_NIST} shows the NIST test results in comparison with single-pixel encryption without scattering. The same Hadamard modulation approach is adopted, where a higher value represents more randomness, and `-' indicates that the data fails the randomness test, which is considered to be structural. We leave out the `Runs' and `LongRuns' factors due to insufficient data lengths. The NIST test shows that the plaintext fails 15 randomness tests, the one-dimensional coupling data modulated by patterns passes four tests, and the scattering-modulated one-dimensional coupling data passes eight tests. Compared with the traditional single-pixel encryption method, the scattering-induced encryption reported in this work enhances the information entropy and randomness of the ciphertext by approximately two times, which notably enhances encryption security.

\section{Conclusion and discussion}

In this work, we report a technique for highly compressed optical sensing and encryption based on scattering-induced entropy boosting, where the scattering transformation multiplexes global information into regional fields, thus achieving scattering enhancement by the compact coupling of feature information. To validate the reported technique, we conducted a series of experiments to study the effects of the sampling rate, the size of the FOV, and the scattering strength on the sensing accuracy. We experimentally demonstrated the technique's effectiveness on both single-target classification and multitarget recognition tasks, and our approach reduces measurements by two orders of magnitude. Specifically, we obtained 95.08\% classification accuracy on the MNIST dataset at a sampling rate of 1\% and 96.78\% recognition accuracy on the Chinese license plate dataset at a sampling rate of 5\%.

We took advantage of the entropy boosts induced by scattering with compressive theory, thus achieving high-precision sensing on highly compressed measurements. This is unlike conventional optical imaging systems \cite{Batarseh2018Passive, kang2017high, naik2011single, yanik2004time, lee2016exploiting, katz2014non, freund1988memory, starck2003gray, pajares2004wavelet} that either ignore optical interference, such as scattering, during the data acquisition process or view light interference as a stage of information degradation \cite{fu2020single}and then introduce an additional complicated denoising step into the workflow. We employ scattering to intrinsically multiplex local features from the whole field to a smaller area. By proactively adopting the moderate-intensity scattering transformation, we achieve image-free sensing with notable advancements in highly compressed data reduction and high-precision perception, while excessively strong scattering may fail to make advancements. The technique is of great value in a resource-limited environment and provides a promising solution for high-throughput machine intelligence with low bandwidth, low costs, and strong encryption.

The reported technique is of great practical value, and can be employed for efficient object detection from highly compressed measurements. For instance, the scattering-enhanced strategy can be combined with advanced optical devices, such as LiDAR, to achieve high-precision sensing on long-distance multidimensional measurements \cite{li2021single}. 
By employing the reported scattering-enhanced strategy, it becomes possible to eliminate the unnecessary step of recovering high-definition 3D targets. We could extract useful features such as position information directly from the detected measurement, leading to an even more efficient detection process. In addition, the technique has the potential to greatly improve the speed and accuracy of data acquisition, and has the potential to significantly improve the performance of various computer vision applications, including motion analysis, object tracking, pose estimation, and so on. By combining it with various image enhancement, image translation, and other advanced deep learning models, it becomes possible to extract high-level features from data and perform complex visual tasks with greater accuracy and reliability. 

In addition, based on the combination of scattering and illuminating light modulation, we innovatively extended the idea to the field of information encryption, thus achieving high-security encryption for highly compressed measurements. Compared with conventional systems, scattering diffusers are less expensive and have lower requirements for the operating environment and the supporting hardware platform. Moreover, the continuous modulated scattering light can contribute to higher sensing precision, as it is completely dependent on the detector rather than on a series of modulation equipment, which could accomplish large-scale modulation and high-precision sensing \cite{ozcan2016lensless}. More importantly, the experiments carried out in this work reveal that scattering disturbs the optical information in the spatial domain, although the information uncertainty, namely, the information entropy, is boosted during the encryption process. Accordingly, the scattering encryption algorithm realizes minimized information compression. In this work, we combined two high-performing information encryption methods with ultrahigh security. Furthermore, we explored contexts with extremely low sampling rates, where the encrypted information still retains adequate characteristic information of the signal, which will open up an avenue to deepen the investigation of optical encryption.

This technique has great potential for further development. Fundamentally, the existing decoding networks may not be that efficient when dealing with targets of different scattering intensities, as the decoding network needs to be trained respectively. Therefore, we propose to adopt the variational network\cite{yue2019variational} for efficient sensing regardless of the diversity of the scattering mediums, thus enhancing the generalization performance of the reported method. Also, we suggest adopting advanced deep learning techniques such as transfer learning to further improve the efficiency and generalization of the technique.
Additionally, the decoding networks adopted in the current work are computationally optimized for various sensing tasks, and better performance will be attained by associating them with the physical characteristics of scattering via model-driven methodologies.
Moreover, the current work employs a DMD for light modulation, which could lead to restricted imaging rates due to its dependency on programmable spatial light modulators. To get around this restriction, we suggest implementing a faster modulation method that relies on cyclic patterns coded onto a spinning mask\cite{hahamovich2021single}. This technique can be combined with scattering transformation to enable real-time high-precision sensing of dynamic objects, which has significant practical implications.
Furthermore, we carried out experiments on target modulation and detection at grayscale levels, while in further work, the scattering-enhanced technique can be applied to detect a wide spectrum of colors. 
We can extract higher feature information and improve sensing precision by implementing scattering modulations in more abundant spectrums.

\subsection{Author Contributions}
X.Z. constructed the overall idea, designed the experiment, and prepared the manuscript. X.C., D.L., R.Y., and Y.Z performed the algorithm study and data analysis. L.B. supervised the project. All the authors participated in discussions during the project.

\subsection{Funding}
National Natural Science Foundation of China (Nos. 62322502, 62131003).

\subsection{Acknowledgment}
A patent with the patent number CN112200264B (G06K9/62) has been granted in regard to this work.

\bibliographystyle{IEEEtran}

\end{document}